# AD-CDO: A Lightweight Ontology for Representing Eligibility Criteria in Alzheimer's Disease Clinical Trials

Zenan Sun, MS,[1] Rashmie Abeysinghe, PhD,[2] Xiaojin Li, PhD,[2] Xinyue Hu, MS,[4] Licong Cui, PhD,[1] Guo-Qiang Zhang, PhD,[2] Jiang Bian, PhD,[3] Cui Tao, PhD[4]

[1]McWilliams School of Biomedical Informatics, University of Texas Health Science Center at Houston, Houston, TX, USA

[2]Department of Neurology, McGovern Medical School, University of Texas Health Science Center at Houston, Houston, TX, USA

[3]Department of Biostatistics and Health Data Science, Indiana University School of Medicine, Indianapolis, IN, USA

[4]Department of Artificial Intelligence and Informatics, Mayo Clinic in Florida, Jacksonville, FL, USA

**Corresponding Author:**

Cui Tao, PhD, Department of Artificial Intelligence and Informatics, Mayo Clinic in Florida, 4500 San Pablo Rd, Jacksonville, FL 32224, USA; Phone: 904-953-0255; e-mail: Tao.Cui@mayo.edu

Word Count: 3979

# ABSTRACT

## Objective
This study introduces the Alzheimer's Disease Common Data Element Ontology for Clinical Trials (AD-CDO), a lightweight, semantically enriched ontology designed to represent and standardize key eligibility criteria concepts in Alzheimer's disease (AD) clinical trials.

## Materials and Methods
We extracted high-frequency concepts from more than 1,500 AD clinical trials on ClinicalTrials.gov and organized them into seven semantic categories: Disease, Medication, Diagnostic Test, Procedure, Social Determinants of Health, Rating Criteria, and Fertility. Each concept was annotated with standard biomedical vocabularies, including the UMLS, OMOP Standardized Vocabularies, DrugBank, NDC, and NLM VSAC value sets. To balance coverage and manageability, we applied the Jenks Natural Breaks method to identify an optimal set of representative concepts.

## Results
The optimized AD-CDO achieved over 63% coverage of extracted trial concepts while maintaining interpretability and compactness. The ontology effectively captured the most frequent and clinically meaningful entities used in AD eligibility criteria. We demonstrated AD-CDO's practical utility through two use cases: (a) an ontology-driven trial simulation system for formal modeling and virtual execution of clinical trials, and (b) an entity normalization task mapping raw clinical text to ontology-aligned terms, enabling consistency and integration with EHR data.

## Discussion
AD-CDO bridges the gap between broad biomedical ontologies and task-specific trial modeling needs. It supports multiple downstream applications, including phenotyping algorithm development, cohort identification, and structured data integration.

## Conclusion
By harmonizing essential eligibility entities and aligning them with standardized vocabularies, AD-CDO provides a versatile foundation for ontology-driven AD clinical trial research.

# INTRODUCTION

## Background and Significance

Alzheimer's Disease (AD) is a progressive neurodegenerative disorder characterized by cognitive decline, memory loss, and deterioration in daily functioning. With the global population aging rapidly, the burden of AD is impacting patients, caregivers, healthcare systems, and the broader research community [1–4]. Despite substantial research efforts and numerous clinical trials, effective treatments for AD remain lacking [5–7]. Clinical trials play a critical role in advancing AD treatment by rigorously testing new therapies for their safety and effectiveness [8,9]. These trials provide eligible participants with early access to promising treatments along with specialized medical care and monitoring. To ensure the accuracy of results, clinical trials have specific eligibility criteria to select participants. However, these eligibility criteria are often presented in unstructured free text, making them difficult to parse, standardize, and reuse. Efforts to extract structured elements from eligibility criteria using natural language processing (NLP) techniques have shown promise, but the extracted elements often lack normalization to standardized terminologies [10,11]. Without semantic normalization, linking eligibility criteria across studies or identifying common inclusion and exclusion patterns becomes challenging.

To address this issue, ontologies have become an increasingly useful tool in biomedical informatics [12–14]. Ontology is a formal, hierarchical representation of knowledge that provides standardized vocabularies for specific domains. It has proven effective in biomedical informatics for tasks such as data integration, semantic annotation, cohort discovery, and phenotyping [15–19]. Well-known biomedical ontologies such as Systematized Nomenclature of Medicine – Clinical Terms (SNOMED CT) [20], the Unified Medical Language System (UMLS) [21], and the Disease Ontology (DO) [22] have enabled significant advancements in clinical informatics by providing reusable, interoperable knowledge structures. However, these resources are typically broad in scope and not optimized for the task-specific modeling of clinical trial eligibility criteria, especially within a disease-specific context such as AD.

Several ontologies have been developed to support AD research, but they vary widely in scope, granularity, and applicability. Most focus on broader research goals such as biomarker discovery, disease mechanisms, or clinical care, rather than addressing the specific needs of clinical trial eligibility criteria. For example, the Alzheimer's Disease Ontology (ADO) provides a comprehensive framework of 1,963 classes and 39,894 axioms covering pathology, genetics, and clinical manifestations [22]. While useful for translational research and data integration, ADO is not optimized for trial-level concepts such as eligibility criteria, interventions, or routine clinical assessments. Its breadth can obscure the frequent and high-impact data elements critical for trial modeling. The Alzheimer Disease Map Ontology (ADMO) focuses on systems biology and molecular mechanisms, representing over 10,000 classes of biological pathways, molecular interactions, and pathophysiological processes [23]. Although valuable for mechanistic modeling, its lack of patient-centric entities and trial-specific constructs limits its use for protocol formalization or cohort selection. The Bilingual Ontology of Alzheimer's Disease and Related Diseases (ONTOAD) aligns with SNOMED CT, UMLS, and Logical Observation Identifiers Names and Codes (LOINC) in English and French to support electronic health record (EHR) integration [24]. However, it has not been updated since 2013 and lacks the granularity needed for eligibility modeling. The European Medical Information Framework for Alzheimer's Disease (EMIF-AD) integrates clinical, genetic, imaging, and biomarker data from European cohorts

[25]. While it supports longitudinal analysis and cross-cohort comparisons, it primarily targets population-level modeling rather than formal representation of trial eligibility criteria. These ontologies provide rich biomedical content but are not designed for the task-specific modeling of eligibility criteria, which often rely on concise, recurrent concepts such as lab tests, diagnoses, prior procedures, and rating scale thresholds. These are inconsistently represented or embedded under broad categories.

To address these challenges, there is a need for a computable, semantically enriched ontology that captures the key concepts and structures of AD clinical trial eligibility criteria and aligns them with standardized biomedical vocabularies. Such a resource would not only enhance interoperability and reusability but also provide the foundation for computational tools that can support clinical trial applications.

## Objective

In this paper, we introduce the Alzheimer's Disease Common Data Entity Ontology for Clinical Trial (AD-CDO), a lightweight ontology designed specifically for representing the most essential eligibility criteria concepts in AD clinical trials. Our contribution can be summarized into three core components:

1. **Ontology Construction:** We extracted high-frequency concepts from over 1,500 AD clinical trials on ClinicalTrials.gov, organizing them into seven clinically meaningful semantic categories: Disease, Medication, Diagnostic Test, Procedure, Social Determinants of Health (SDoH), Rating Criteria, and Fertility. Each category represents a core dimension of eligibility criteria logic. To enhance interoperability, concepts are enriched with value sets and cross-references to standard vocabularies such as UMLS concept unique identifier (CUI), standard concept IDs of the Observational Medical Outcomes Partnership (OMOP) Standardized Vocabularies, DrugBank category, National Drug Code (NDC), and National Library of Medicine Value Set Authority Center (NLM VSAC) valuesets. This structure allows AD-CDO to function as both a knowledge model and a foundation for phenotype algorithm generation and cohort identification.
2. **Ontology Evaluation:** We applied a data-driven strategy using Jenks Natural Breaks (JNB) to determine an optimal set of concepts per category that balances coverage of trial data and ontology manageability. This ensures that AD-CDO remains lightweight while representing the most essential and commonly used eligibility entities.
3. **Real-World Applications:** We demonstrate AD-CDO's practical utility through two use cases: (a) an ontology-driven trial simulation system that enables formal modeling and virtual execution of clinical trials using Temporal Event Logic (TEL) [26], and (b) an entity normalization pipeline that maps raw clinical text in trial protocols to standardized, ontology-aligned terms, improving consistency and enabling integration with EHR data.

# MATERIALS AND METHODS

The AD-CDO was developed to model standardized AD clinical trial elements, with a focus on structured representation of eligibility criteria concepts. The ontology was manually constructed using Protégé and represented in the Web Ontology Language (OWL) [27]. As illustrated in Figure 1, the overall development workflow consists of three major components: (A) data sources and preparation, (B) ontology construction, (C) ontology evaluation and refinement. We then demonstrated the AD-CDO in two real-world applications: a trial simulation system and entity normalization use case.

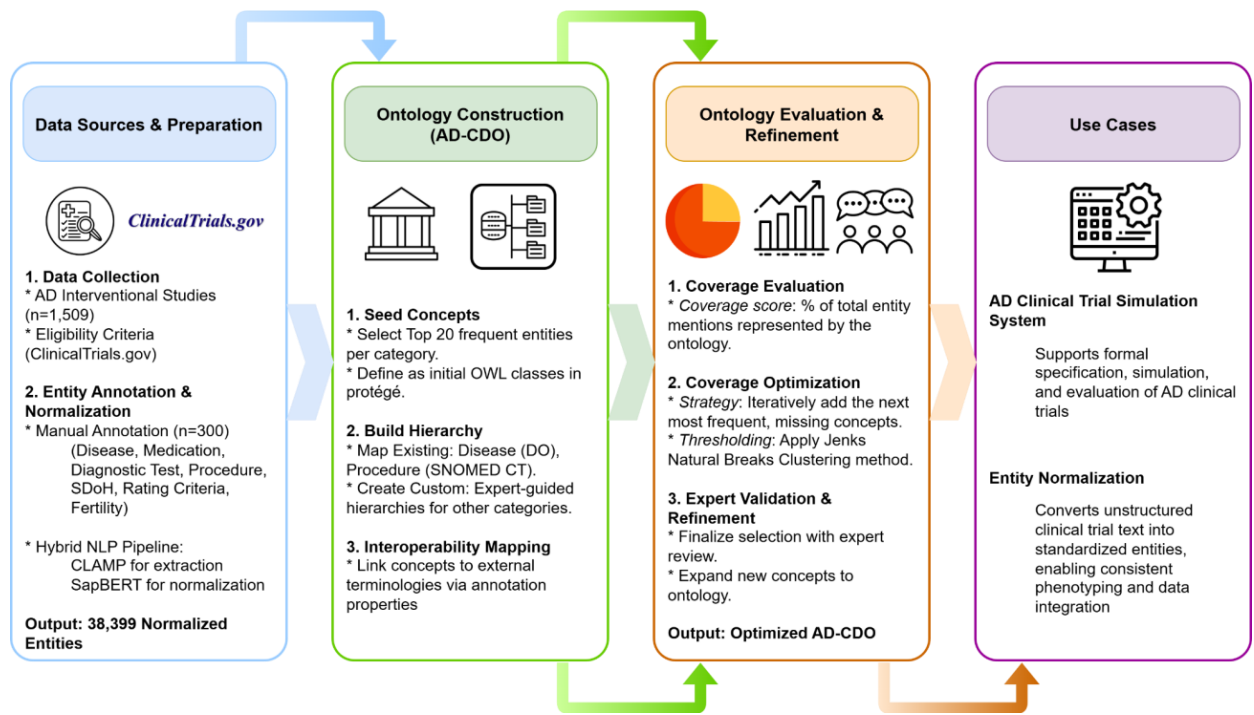

Figure 1. Overview of the AD-CDO development pipeline: preparation, construction, evaluation, and use cases.

## Data Sources And Preparation

To construct a representative and semantically rich ontology, we began by collecting and preprocessing eligibility criteria data from ClinicalTrials.gov, specifically focusing on interventional studies related to AD (n = 1,509). In our previous work [10], we have manually annotated a corpus of 300 AD clinical trials, identifying and categorizing entities into seven predefined semantic types: Disease, Medication, Diagnostic Test, Procedure, SDoH, Rating Criteria, and Fertility. These categories were selected based on their prevalence and relevance within the eligibility criteria and aligned with common clinical data element frameworks.

To expand the coverage beyond the manually annotated set, we applied a hybrid NLP pipeline to the remaining AD trials. First, the Clinical Language Annotation, Modeling, and Processing (CLAMP) toolkit [28] was used to perform named entity recognition (NER), extracting candidate biomedical concepts from free-text eligibility criteria. Then, we employed SapBERT [29], a transformer-based biomedical entity linking model, to normalize the extracted entities to concepts to the UMLS standardized vocabulary. This process yielded a total of 38,399 normalized entity mentions from the full set of AD trials.

The resulting dataset formed the foundation for selecting representative concepts to include in the initial version of AD-CDO, ensuring that the ontology reflects real-world trial data and supports interoperability with existing biomedical terminologies.

**Ontology Construction**

The ontology was constructed following a stepwise workflow. First, for each of the seven predefined semantic categories, we identified the top 20 most frequently occurring normalized entities from the extracted clinical trial data to serve as initial seed concepts. These seed entities formed the basis for defining the core classes and subclasses of AD-CDO, with each concept formally represented as an OWL class.

Next, we built the hierarchical structure for each category using a combination of external ontology mappings and expert-guided manual curation:

- *Disease:* Concepts were mapped to the DO, inheriting its superclass relationships to capture established disease hierarchies.
- *Procedure:* Concepts were aligned with the SNOMED CT hierarchy, using its clinically validated procedural taxonomy.
- *Medication, Diagnostic Test, SDoH, Rating Criteria, Fertility:* Custom hierarchies were created using term frequency, clinical guidelines, and domain expertise to ensure practical relevance and comprehensive coverage.

This hybrid approach allowed us to construct a semantically coherent ontology that balances interoperability with the flexibility to represent AD-specific concepts.

To ensure interoperability and facilitate integration with widely adopted biomedical standards, we manually mapped each curated concept to relevant external terminologies. For every concept included in the ontology, we identified equivalent or semantically aligned entries across multiple standardized biomedical resources. These mappings were implemented in Protégé using custom annotation properties, defined as follows:

- *hasAthenaID:* maps the concept to its corresponding standardized concept identifier in the OMOP Standardized Vocabularies, ensuring alignment with OMOP common data model (CDM) standards.
- *hasUMLS:* links the concept to its normalized UMLS CUI, providing semantic grounding and compatibility with a broad range of biomedical tools and datasets.
- *hasBrandName:* used for medications that have multiple branded formulations, offering additional granularity for drug-related concepts.
- *hasDrugBankCategory:* links medication entities to their respective drug categories in the DrugBank database, enhancing pharmacological context.
- *hasNDC:* provides associated National Drug Codes for medication concepts, supporting integration with pharmacy and prescription datasets.

- *hasValuesets:* associates concepts with relevant valuesets from the NLM VSAC, enabling alignment with clinical quality measures and EHR systems.

For manual mapping to external vocabularies, we ensured that each entity was linked to the most appropriate standardized terminology based on its semantic category. Specifically, for mapping to OMOP Standardized Vocabularies, Disease entities were linked by SNOMED CT concepts, Procedure entities were linked by both SNOMED CT concepts and Current Procedural Terminology (CPT) codes, Medication entities were linked by RxNorm and ATC codes, and Diagnostic Test entities were linked by LOINC codes. Medication entities were additionally annotated with DrugBank categories and NDC codes. Furthermore, all entities were mapped to UMLS CUI to provide semantic grounding and interoperability with biomedical datasets. In addition, we manually annotated curated value sets from the NLM VSAC. These value sets provide comprehensive and semantically coherent lists of codes for each clinical concept and enable consistent representation across heterogeneous data sources. By linking concepts to NLM VSAC valuesets, AD-CDO supports computable phenotyping, cohort definition, and integration with EHR systems, ensuring that trial eligibility criteria are standardized and can be mapped to real-world patient data in an interoperable manner. All these mappings were validated through manual review by two domain experts. Where an exact match was unavailable, the closest semantically aligned concept was selected.

## Ontology Evaluation and Refinement

To assess the quality and practical utility of AD-CDO, we conducted a two-phase evaluation process consisting of coverage evaluation and coverage optimization.

*Coverage Evaluation:* Coverage score was calculated as the proportion of normalized entity mentions from all extracted entities (n = 38,399) that were represented in AD-CDO. This metric indicates how completely AD-CDO represents the semantic content of real-world AD clinical trial eligibility criteria. High coverage ensures that key clinical concepts found in trial eligibility criteria are captured in the ontology. Therefore, it will support accurate mapping, consistent interpretation, and downstream applications such as cohort identification and automated screening.

*Coverage Optimization:* Excessively large ontologies can become cumbersome and redundant, while smaller ontologies may omit important concepts. To enhance representational coverage without unnecessary expansion, we implemented an iterative enrichment strategy. In each iteration, the next most frequent yet unrepresented concept was added to its corresponding semantic class, and the new coverage score was recalculated.

Our goal was to identify an optimal threshold beyond which additional concepts yielded minimal improvement. To determine this threshold, we applied the JNB optimization method, originally developed for geospatial classification [30,31]. The method partitions continuous data (e.g., a series of coverage scores) into clusters that minimize intra-class variance and maximize inter-class separation. We configured JNB to separate the results into two clusters:

- A "core cluster" of high-impact additions that substantially improved coverage, and
- A "long-tail cluster" of additions with limited benefit.

The break point between these clusters identified the optimal threshold, balancing coverage performance and ontology compactness. Concepts beyond this point were considered less essential for representing common data elements in AD trials.

After identifying the optimal threshold, we expanded AD-CDO with the selected concepts and updated the class hierarchies. Furthermore, our domain experts also reviewed candidate concepts and agreed on a set of additional terms to include based on their clinical relevance. All newly incorporated concepts were manually verified to ensure semantic accuracy, consistency with class definitions, and correct mappings to external vocabularies. This combined data-driven and expert-guided strategy produced a lightweight yet expressive ontology, optimized for real-world AD clinical trial representation and downstream informatics applications. Following these refinements, a final coverage score was calculated to quantify the representational completeness of the optimized AD-CDO.

## RESULTS

### AD-CDO Overall Structure

The current version of AD-CDO comprises 293 classes, 305 logical axioms, 303 declaration axioms, and 6 annotation properties. Figure 2 illustrates the high-level architecture of AD-CDO in Protégé, showcasing its organization into seven main categories.

The Disease category adopts the top two levels of the Disease Ontology hierarchy to maintain semantic alignment with established biomedical vocabularies. Extracted disease-related elements are then mapped to corresponding leaf nodes to ensure granularity and consistency. The Procedure category follows the hierarchical framework of SNOMED CT and includes nine sub-classes for standardized representation of clinical interventions. The Diagnostic Test, Medication, Rating Criteria, and SDoH categories contain eleven, six, four, and five sub-classes, respectively. Due to the narrow scope and limited subclass structure, Fertility is represented as a standalone class. Overall, the classification structure combines expert knowledge with ontology standards to ensure consistency and interoperability.

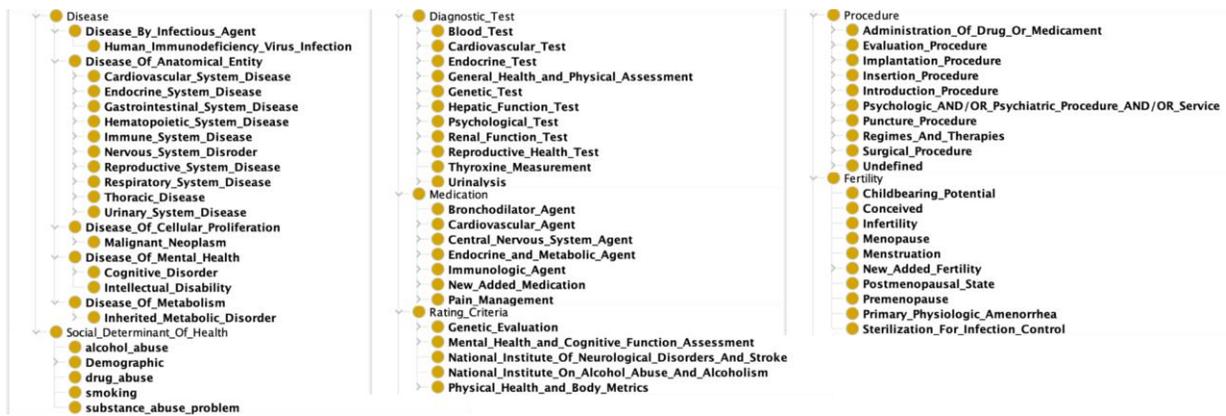

Figure 2. The seven main categories and direct subclasses of AD-CDO in Protégé

Figure 3 presents an example of annotation properties for the concept Epilepsy. This example illustrates how AD-CDO links Epilepsy to multiple standardized resources, creates a rich semantic representation that supports interoperability and computability, and serves as a phenotyping library. The hasUMLS property assigns a universal identifier from the UMLS, unifying synonyms and variants across vocabularies. This ensures semantic grounding and compatibility with a wide range of biomedical tools and datasets. The hasAthenaID property enables direct alignment with the OMOP CDM via its standardized vocabularies. Through this mapping, Epilepsy is linked to a standardized SNOMED CT concept and inherits relationships to related medical concepts, allowing AD-CDO to be applied seamlessly in large-scale observational research and clinical databases. The hasValuesets property plays a key role by aggregating all clinically relevant codes representing a concept across multiple standardized resources, thereby enhancing interoperability and semantic richness. For Epilepsy, this means International Classification of Diseases (ICD), SNOMED CT, and other diagnostic codes are curated into a single, validated list. By linking AD-CDO concepts to NLM VSAC value sets, eligibility criteria can be consistently interpreted across heterogeneous data sources and translated into executable queries for EHR or claims databases. With these annotations, AD-CDO enhances semantic interoperability, facilitates entity normalization across systems, and enables downstream applications such as EHR integration, computable phenotyping, and automated eligibility reasoning.

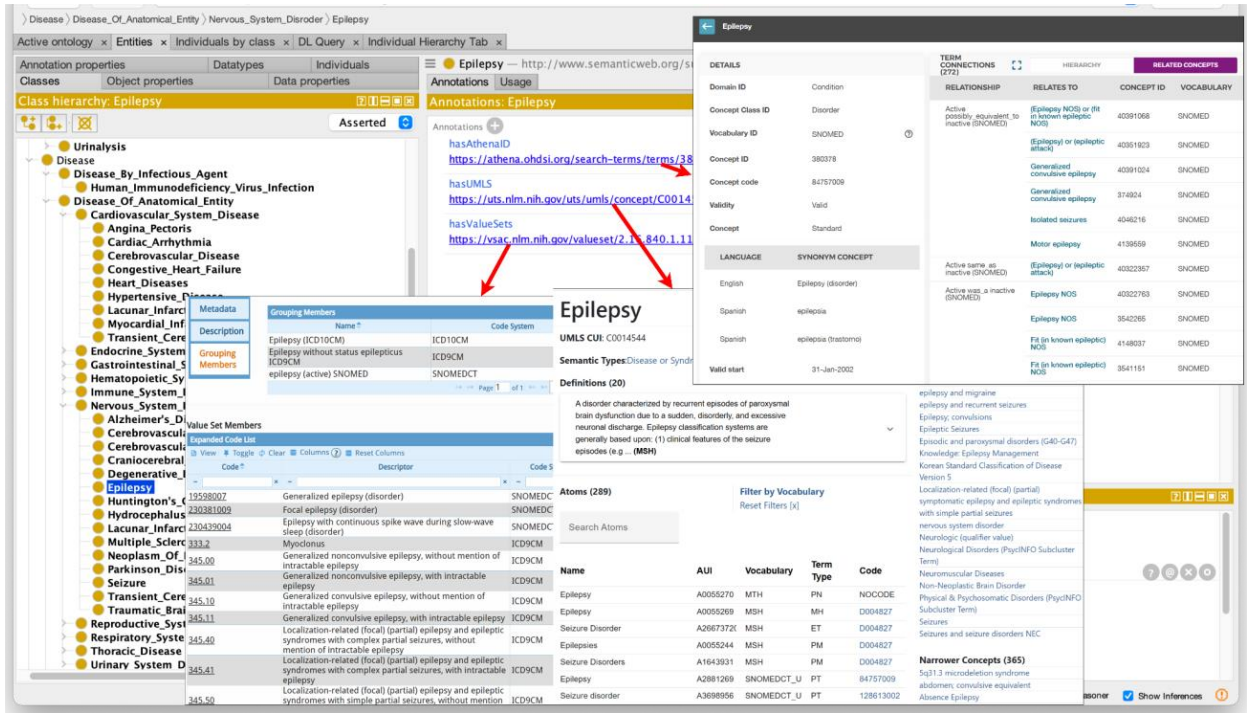

Figure 3. AD-CDO Annotation Property Example

In the following section, we report the evaluation result of the AD-CDO using the coverage analysis, which measures the extent to which the ontology captures real-world clinical trial entities.

## Coverage Analysis

The initial version of AD-CDO was built using the 20 most frequent concepts from each of the seven semantic classes. This version achieved a coverage score of 0.5025, indicating that 50.25% of all extracted entity mentions were represented in the ontology. This result shows that a small set of high-frequency concepts can capture much of the key information in AD trial eligibility criteria, demonstrating the efficiency of frequency-based selection in building a compact ontology.

To enhance coverage, we applied a data-driven selection strategy using the JNB method. As shown in Figure 4, the method identified p = 29 as the optimal threshold between high- and low-impact additions, yielding a coverage score of 0.6285. By expanding the ontology to 49 (20 + 29) concepts per class, approximately 63% of trial entities were captured. For categories with fewer than 49 frequent concepts, all relevant entities were included. Because SDoH entities were highly heterogeneous and variable in granularity, no additional concepts were added beyond the core set.

Finally, after incorporating manually reviewed candidate concepts suggested by domain experts, the final coverage score of AD-CDO increased to 0.6341. This optimized configuration strikes a balance between representational comprehensiveness and compactness, resulting in a lightweight yet expressive ontology tailored for downstream clinical trial modeling and informatics applications.

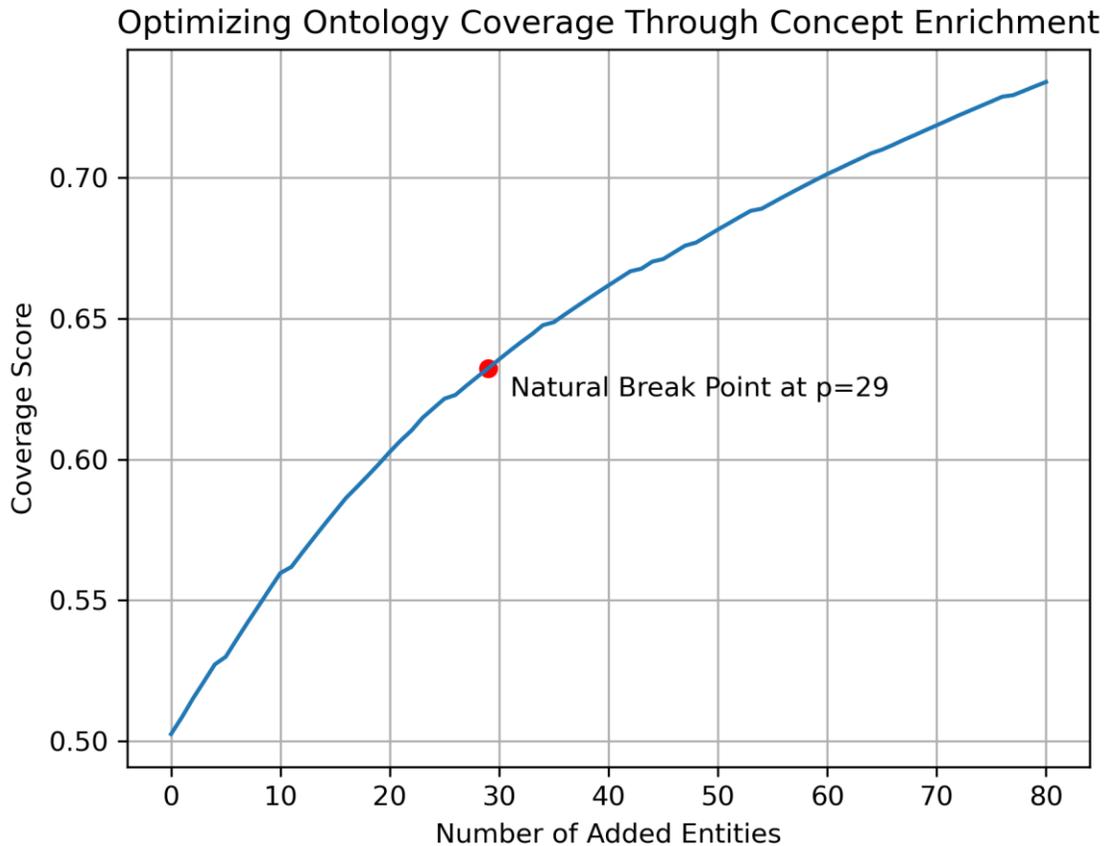

Figure 4. Optimizing Ontology Coverage Through Concept Enrichment

## Case Study

In addition to its structural contributions, AD-CDO has been applied in two real-world use cases that demonstrate its practical utility: (a) an ontology-driven clinical trial simulation system that enables the configuration, execution, and feasibility assessment of trial protocols using real-world EHR data; and (b) an entity normalization example that transforms unstructured eligibility criteria into standardized, computable concepts aligned with external biomedical vocabularies.

*AD Clinical Trial Simulation System:* To demonstrate the real-world utility of the AD-CDO, we integrated it into the AD Clinical Trial Simulation System [32], a logic-based platform designed to support the formal specification, simulation, and evaluation of AD clinical trials using observational real-world data.

The system allows researchers to design and simulate virtual trials through a web-based interface. Built upon AD-CDO, it provides a semantically consistent and computable framework for representing key trial components, including eligibility criteria, interventions, outcomes, temporal dependencies, and index events. By translating free-text trial protocols into formal computational models, the platform bridges the gap between narrative designs and executable representations. It employs TEL to capture complex time-dependent relationships among clinical events, ensuring accurate modeling of eligibility windows, treatment phases, follow-up intervals, and outcomes.

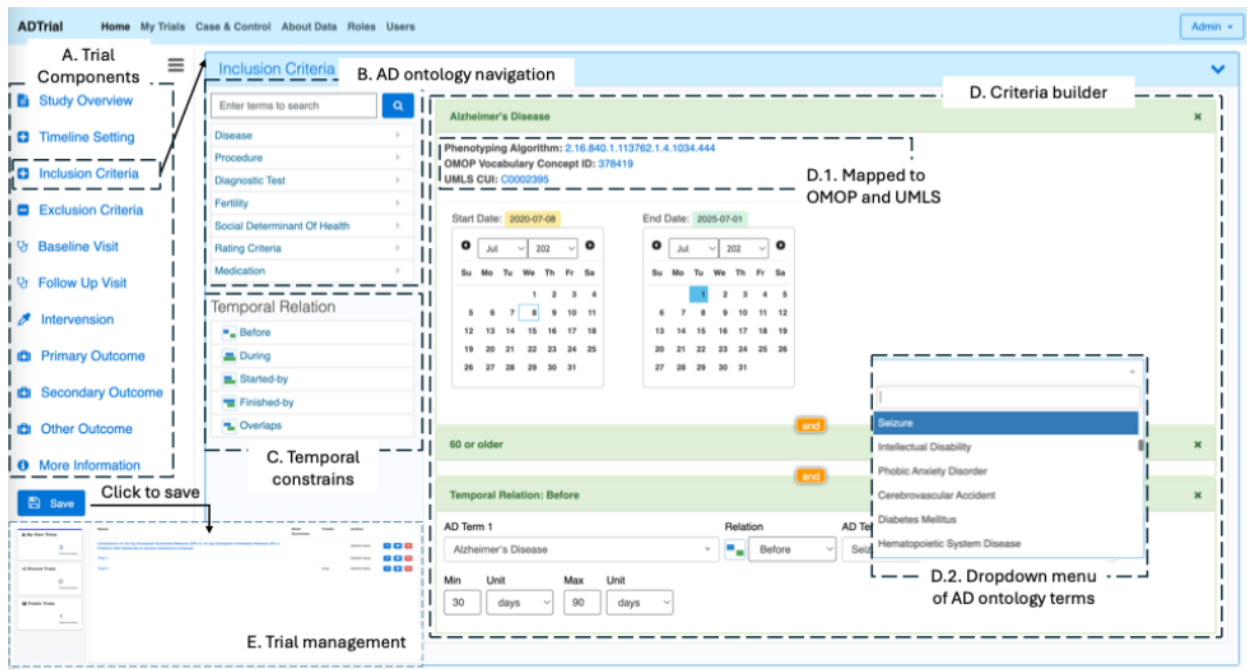

Figure 5. The interface of the clinical trial simulation system using AD-CDO

AD-CDO serves as the system's controlled vocabulary, guiding the semantic configuration of all trial elements. Ontology concepts are integrated throughout the interface to ensure alignment and consistency. The interface includes four main components:

- Ontology-Aligned Criteria Builder: Enables users to specify inclusion and exclusion criteria by selecting standardized concepts from AD-CDO's seven semantic categories (Figure 5A).
- Ontology Navigation Interface: Supports exploration and selection of relevant ontology concepts for transparent and reusable trial definitions (Figure 5B).
- Temporal Configuration Tools: Allow users to define time-dependent logic using TEL constructs anchored to specific index events (Figure 5C–D).
- Trial Management and Simulation Interface: Facilitates reviewing, editing, and exporting trials. Configured trials are compiled into TEL expressions and executed against real-world EHR data to simulate recruitment, assess feasibility, and compare study designs (Figure 5E).

The integration of AD-CDO into the AD Clinical Trial Simulation System illustrates a high-impact use case of ontology-based modeling. By grounding complex trial logic in a structured semantic framework, the system empowers researchers to design and simulate protocol-driven studies in a manner that is transparent, reusable, and computationally robust.

*Normalization Use Case:* To illustrate this capability, we applied AD-CDO to answer a representative question such as *"Which clinical codes are typically collected for patients with Alzheimer's disease?"* as shown in Figure 6. We first extract key terms from the query; in this example, "AD" is normalized through AD-CDO to the formally defined ontology concept "Alzheimer's Disease." This concept is further categorized into its semantic class (e.g., Disease) and enriched with external identifiers such as UMLS CUIs, OMOP Standardized Vocabularies standard concept IDs, and curated ICD value sets from NLM VSAC.

Beyond entity normalization, AD-CDO also provides associated value sets relevant to AD, including medications used for treatment (e.g., cholinesterase inhibitors) and cognitive assessments with their corresponding medical codes. By linking diseases to their related clinical entities, AD-CDO functions not only as an ontology but also as a phenotyping library that captures disease-defining features across multiple clinical dimensions.

By resolving synonymy, abbreviation, and free-text variation, AD-CDO supports reliable entity normalization that facilitates cohort identification, computable phenotyping, and automated trial screening. Moreover, the process transforms raw and potentially ambiguous text into structured, computable, and interoperable representations, thereby enhancing semantic clarity and establishing a foundation for automated reasoning, cross-study harmonization, and AI-driven clinical research.

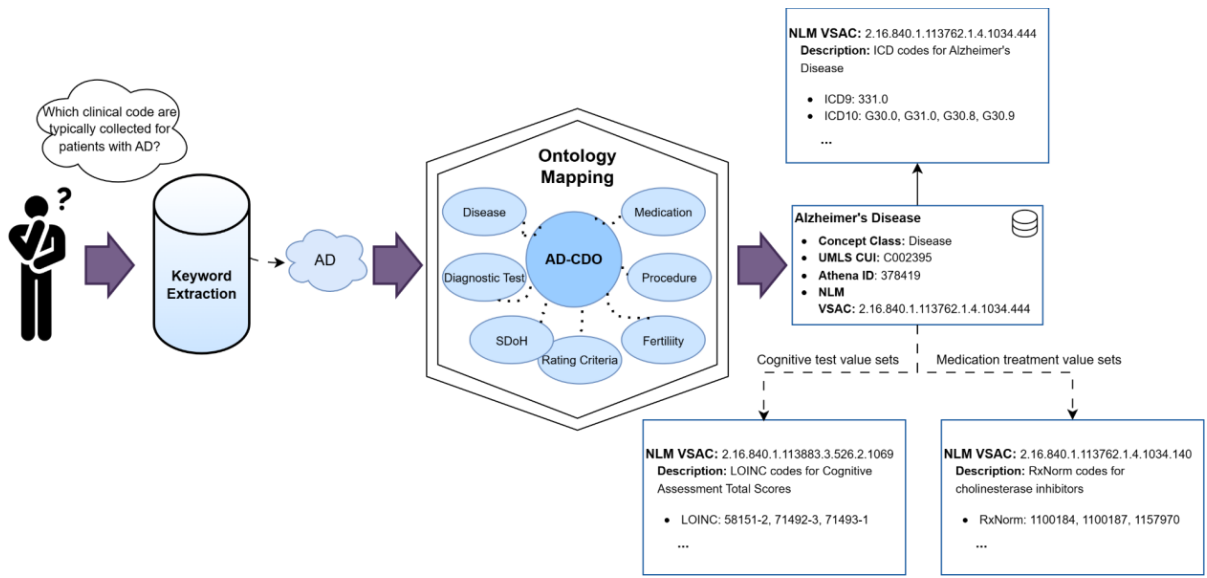

Figure 6. AD-CDO for the entity normalization pipeline example

# DISCUSSION

AD-CDO is a domain-specific ontology designed to formalize eligibility criteria in AD clinical trials. Unlike existing Alzheimer's ontologies, which predominantly emphasize molecular mechanisms or broad biomedical domains, AD-CDO focuses on the most frequently occurring eligibility-related entities in AD trials. Its vocabulary is structured into seven semantic categories, which provides targeted yet comprehensive coverage of trial-relevant concepts. This focused scope fills a critical gap by addressing the practical needs of clinical trial informatics rather than duplicating broad biomedical knowledge bases.

A key strength of AD-CDO is the integration of ontology engineering with empirical evidence. Concepts were prioritized using frequency analysis, while JNB optimization provided a principled mechanism for selecting representative concepts per category. This methodological rigor ensures that the ontology balances breadth of coverage with maintainability. Furthermore, the manual mapping of all concepts to external vocabularies, such as UMLS CUIs, standard concept IDs of the OMOP Standardized Vocabularies, DrugBank categories, NDC codes, and NLM VSAC value sets, supports precise semantic interoperability. The inclusion of value sets is particularly important, as it facilitates harmonized use of clinical terminologies across heterogeneous data sources, which is critical for consistent phenotyping. From a practical perspective, AD-CDO provides several advantages. First, it supports phenotyping pipelines, enabling unstructured eligibility criteria to be transformed into computable, standardized representations suitable for cohort identification. Second, the ontology's value set mappings enhance integration with EHR data and real-world data platforms, bridging the gap between trial eligibility design and patient-level datasets. Third, its lightweight yet high-coverage design makes it suitable for scalable applications such as trial feasibility assessment, cohort discovery, and clinical trial simulation.

While AD-CDO provides a structured and semantically aligned representation of common data elements used in Alzheimer's Disease clinical trials, several limitations remain. First, the current version of the ontology focuses solely on class-level organization and concept annotation, and does not yet incorporate formal relationships between concepts. As a result, it cannot capture richer semantics, such as "Medication X has an adverse interaction with Medication Y," which are essential for enabling reasoning and advanced query capabilities. Second, although the ontology includes mappings to multiple external vocabularies, the mapping process was conducted manually, which is both labor-intensive and may introduce inconsistencies. Incorporating automated or semi-automated mapping techniques in future versions could improve scalability, reduce manual effort, and enhance consistency across concepts. Third, the ontology's current coverage and structure are derived from AD clinical trial eligibility criteria only, and thus may not fully represent other AD-related domains such as biomarkers, imaging data, or longitudinal clinical outcomes. While this narrow scope is effective for modeling eligibility criteria, it limits the ontology's applicability to broader domains in AD.

# CONCLUSION

In this study, we developed AD-CDO to address the absence of standardized, semantically rich representations of eligibility criteria in AD clinical research. Drawing from more than 1,500 AD trials, we extracted and normalized high-frequency concepts to construct a lightweight yet high-coverage ontology organized into seven semantic categories: Disease, Medication, Diagnostic Test, Procedure, Rating Criteria, SDoH, and Fertility. Each concept was manually mapped to external biomedical vocabularies, including UMLS CUI, standard concept IDs of the OMOP Standardized Vocabularies, DrugBank Categories, NDC, and NLM VSAC value sets. This mapping not only ensures interoperability but also aligns AD-CDO with standardized value sets that can support consistent phenotyping and integration with EHR and real-world data sources. To demonstrate practical utility, we presented two use cases: a clinical trial simulation platform and an entity normalization example. AD-CDO thus provides a foundational resource for eligibility modeling, cohort identification, phenotyping, and protocol standardization in AD research. Current limitations include the absence of explicitly defined relationships and reliance on manual mappings. Future work will focus on expanding coverage, introducing automated mapping strategies, and integrating relational semantics to enable more advanced reasoning and downstream applications.

# COMPETING INTERESTS STATEMENT

The authors declare no competing interests.

# FUNDING STATEMENT

This research was supported by NIH grants under Award Numbers R01AG084236, R01AG083039, and RF1AG072799.

# AUTHOR CONTRIBUTIONS

ZS: Conceptualized the study, curated and interpreted data, constructed and refined the ontology, and drafted the manuscript.

RA: Contributed to ontology construction and refinement.

XL and XH: Contributed to ontology refinement and design of use cases.

LC, GZ, JB: Contributed to funding acquisition and manuscript review and editing.

CT: Contributed to conceptualization, funding acquisition, supervision, and manuscript review and editing.